\documentclass{llncs}

\usepackage{graphicx}
\usepackage{subfigure}
\usepackage{url}

\begin{document}

\title{The Importance of Skip Connections in Biomedical Image Segmentation}
\titlerunning{The Importance of Skip Connections}  
%
\author{Michal Drozdzal\inst{1,2, \footnotemark[1]}\footnotetext{\footnotemark[1] Equal contribution\\ Accepted for DLMIA 2016} \and Eugene Vorontsov\inst{1,2, \footnotemark[1]} \and Gabriel Chartrand\inst{1,3} \and \\ Samuel Kadoury\inst{2,4} \and Chris Pal\inst{2,5}}
\authorrunning{Michal Drozdzal, Eugene Vorontsov et. al.} 
%
\tocauthor{Michal Drozdzal, Eugene Vorontsov, Gabriel Chartrand, Samuel Kadoury, Chris Pal}
\institute{Imagia Inc., Montr\'{e}al, Canada \\
\email{\{michal, eugene, gabriel\}@imagia.com} \\
\email{\{samuel.kadoury, christopher.pal\}@polymtl.ca }
\and
\'{E}cole Polytechnique de Montr\'{e}al, Montr\'{e}al, Canada\\
\and
Universit\'{e} de Montr\'{e}al, Montr\'{e}al, Canada
\and
CHUM Research Center, Montr\'{e}al, Canada
\and
Montreal Institute for Learning Algorithms, Montr\'{e}al, Canada
}

\maketitle              

\begin{abstract}
In this paper, we study the influence of both long and short skip connections on Fully Convolutional Networks (FCN) for biomedical image segmentation. In standard FCNs, only long skip connections are used to skip features from the contracting path to the expanding path in order to recover spatial information lost during downsampling. We extend FCNs by adding short skip connections, that are similar to the ones introduced in residual networks, in order to build very deep FCNs (of hundreds of layers). A review of the gradient flow confirms that for a very deep FCN it is beneficial to have both long and short skip connections. Finally, we show that a very deep FCN can achieve near-to-state-of-the-art results on the EM dataset without any further post-processing.
\end{abstract}
\section{Introduction}
\label{sec:intro}

Semantic segmentation is an active area of research in medical image analysis. With the introduction of Convolutional Neural Networks (CNN), significant improvements in performance have been achieved in many standard datasets. For example, for the EM ISBI 2012 dataset~\cite{EM_data}, BRATS~\cite{Brats} or MS lesions~\cite{MsLesions}, the top entries are built on CNNs \cite{RonnebergerFB15,0011QCH16,HavaeiDWBCBPJL15,Brosch}.

All these methods are based on Fully Convolutional Networks (FCN)~\cite{long_shelhamer_fcn}. While CNNs are typically realized by a contracting path built from convolutional, pooling and fully connected layers, FCN adds an expanding path built with deconvolutional or unpooling layers. The expanding path recovers spatial information by merging features skipped from the various resolution levels on the contracting path.

Variants of these skip connections are proposed in the literature. In \cite{long_shelhamer_fcn}, upsampled feature maps are summed with feature maps skipped from the contractive path while \cite{RonnebergerFB15} concatenate them and add convolutions and non-linearities between each upsampling step. These skip connections have been shown to help recover the full spatial resolution at the network output, making fully convolutional methods suitable for semantic segmentation. We refer to these skip connections as long skip connections.

Recently, significant network depth has been shown to be helpful for image classification \cite{SzegedyLJSRAEVR14,HeZRS15,HeZR016,RomeroBKCGB14}. The recent results suggest that depth can act as a regularizer \cite{HeZRS15}. However, network depth is limited by the issue of vanishing gradients  when backpropagating the signal across many layers. In \cite{SzegedyLJSRAEVR14}, this problem is addressed with additional levels of supervision, while in \cite{HeZRS15,HeZR016} skip connections are added around non-linearities, thus creating shortcuts through which the gradient can flow uninterrupted allowing parameters to be updated deep in the network. Moreover, \cite{SzegedyIV16} have shown that these skip connections allow for faster convergence during training. We refer to these skip connections as short skip connections.

In this paper, we explore deep, fully convolutional networks for semantic segmentation. We expand FCN by adding short skip connections that allow us to build very deep FCNs. With this setup, we perform an analysis of short and long skip connections on a standard biomedical dataset (EM ISBI 2012 challenge data). We observe that short skip connections speed up the convergence of the learning process; moreover, we show that a very deep architecture with a relatively small number of parameters can reach near-state-of-the-art performance on this dataset. Thus, the contributions of the paper can be summarized as follows:
%
%
\begin{itemize}
\item We extend Residual Networks to fully convolutional networks for semantic image segmentation (see Section \ref{sec:ResNet}).
\item We show that a very deep network without any post-processing achieves performance comparable to the state of the art on EM data (see Section \ref{sec:em_data}).
\item We show that long and short skip connections are beneficial for convergence of very deep networks (see Section \ref{sec:skip})
%
%
\end{itemize}

\section{Residual network for semantic image segmentation}
\label{sec:ResNet}

\begin{figure}[t!]
\centering
\subfigure[]{\includegraphics[width=0.5\textwidth]{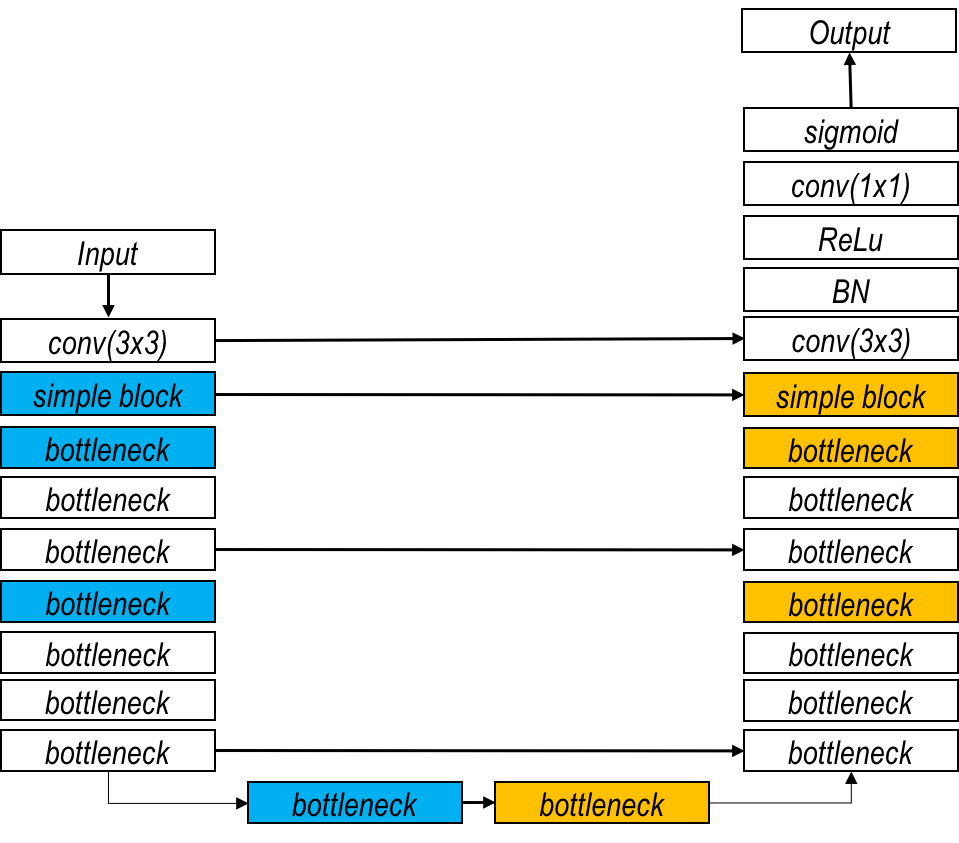}\label{fig:1}}\hfill
\subfigure[]{\includegraphics[width=0.15\textwidth]{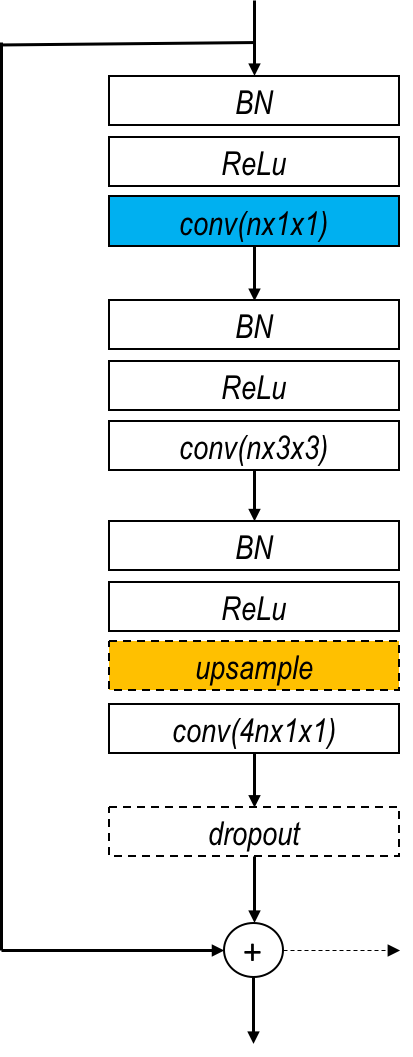}\label{fig:2}}\hfill
\subfigure[]{\includegraphics[width=0.15\textwidth]{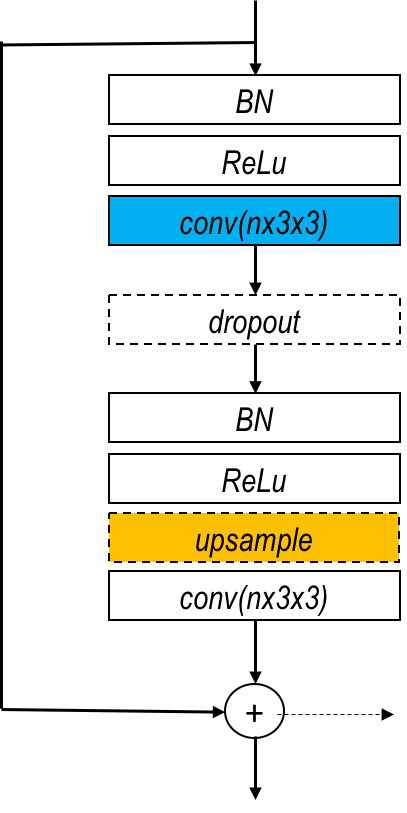}\label{fig:3}}\hfill
\subfigure[]{\includegraphics[width=0.15\textwidth]{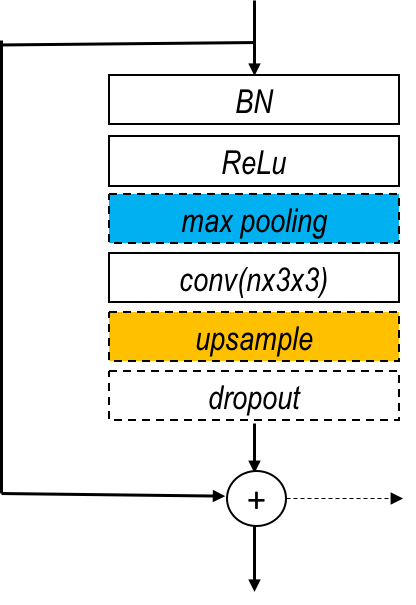}\label{fig:4}}\hfill
\vspace{-.2cm}
\caption{An example of residual network for image segmentation. (a) Residual Network with long skip connections built from bottleneck blocks, (b) bottleneck block, (c) basic block and (d) simple block. Blue color indicates the blocks where an downsampling is optionally performed, yellow color depicts the (optional) upsampling blocks, dashed arrow in figures (b), (c) and (d) indicates possible long skip connections. Note that all blocks (b), (c) and (d) can have a dropout layer (depicted with dashed line rectangle).}
\label{fig:ResUNet}
\vspace{-.5cm}
\end{figure}

Our approach extends Residual Networks \cite{HeZRS15} to segmentation tasks by adding an expanding (upsampling) path (Figure \ref{fig:1}). We perform spatial reduction along the contracting path (left) and expansion along the expanding path (right). As in \cite{long_shelhamer_fcn} and \cite{RonnebergerFB15}, spatial information lost along the contracting path is recovered in the expanding path by skipping equal resolution features from the former to the latter. Similarly to the short skip connections in Residual Networks, we choose to sum the features on the expanding path with those skipped over the long skip connections.

We consider three types of blocks, each containing at least one convolution and activation function: bottleneck, basic block, simple block (Figure \ref{fig:2}-\ref{fig:4}). Each block is capable of performing batch normalization on its inputs as well as spatial downsampling at the input (marked blue; used for the contracting path) and spatial upsampling at the output (marked yellow; for the expanding path). The bottleneck and basic block are based on those introduced in \cite{HeZRS15} which include short skip connections to skip the block input to its output with minimal modification, encouraging the path through the non-linearities to learn a residual representation of the input data. To minimize the modification of the input, we apply no transformations along the short skip connections, except when the number of filters or the spatial resolution needs to be adjusted to match the block output. We use $1\times1$ convolutions to adjust the number of filters but for spatial adjustment we rely on simple decimation or simple repetition of rows and columns of the input so as not to increase the number of parameters. We add an optional dropout layer to all blocks along the residual path.

We experimented with both binary cross-entropy and dice loss functions. Let $o_i \in [0, 1]$ be the $i^{th}$ output of the last network layer passed through a sigmoid non-linearity and let $y_i \in \{0, 1\}$ be the corresponding label. The binary cross-entropy is then defined as follows:

\begin{equation}
L_{bce} = \sum_i y_i \log{o_i} + (1-y_i) \log{(1-o_i)}
\end{equation}

The dice loss is: 
\begin{equation}
L_{Dice} = -\frac{2 \sum_i o_i y_i}{\sum_i o_i + \sum_i y_i}
\end{equation}
We implemented the model in Keras \cite{chollet2015keras} using the Theano backend \cite{theano} and trained it using RMSprop\cite{Tieleman2012} (learning rate $0.001$) with weight decay set to $0.001$. We also experimented with various levels of dropout.


\section{Experiments}
\label{sec:experiments}
In this section, we test the model on electron microscopy (EM)  data \cite{EM_data} (Section \ref{sec:em_data}) and perform an analysis on the importance of the long and short skip connections (Section \ref{sec:skip}).

\subsection{Segmenting EM data}
\label{sec:em_data}

EM training data consist of $30$ images ($512 \times 512$ pixels) assembled from serial section transmission electron microscopy of the Drosophila first instar larva ventral nerve cord. The test set is another set of $30$ images for which labels are not provided. Throughout the experiments, we used $25$ images for training, leaving $5$ images for validation. 

During training, we augmented the input data using random flipping, sheering, rotations, and spline warping. We used the same spline warping strategy as \cite{RonnebergerFB15}. We used full resolution  ($512\times512$) images as input without applying random cropping for data augmentation. For each training run, the model version with the best validation loss was stored and evaluated. The detailed description of the highest performing architecture used in the experiments is shown in Table \ref{tab:arch}.

\begin{table}[t!]
\begin{center}
\footnotesize
\begin{tabular}{||c |c |c |c |c||} 
\hline
Layer name & block type & output resolution & output width & repetition number \\ [0.5ex] 
\hline\hline
Down 1 & conv $3 \times 3$ & $512 \times 512$ & 32 & 1\\
\hline
Down 2 & simple block & $256 \times 256$ & 32 & 1 \\ 
\hline
Down 3 & bottleneck & $128 \times 128$ & 128 & 3\\
\hline
Down 4 & bottleneck & $64 \times 64$ & 256 & 8 \\
\hline
Down 5 & bottleneck & $32 \times 32$ & 512 & 10 \\
\hline
Across & bottleneck & $32\times 32$ & 1024 & 3 \\
\hline
Up 1 & bottleneck & $64 \times 64$ & 512 & 10 \\
\hline
Up 2 & bottleneck & $128 \times 128$ & 256 & 8\\
\hline
Up 3 & bottleneck & $256 \times 256$ & 128 & 3 \\
\hline
Up 4 & simple block & $512 \times 512$ & 32 & 1\\
\hline
Up 5 & conv 3x3 & $512 \times 512$ & 32 & 1 \\
\hline
Classifier & conv $1 \times 1$ & $512 \times 512$ & 1 & 1\\ 
\hline
\end{tabular}
\end{center}
\caption{Detailed model architecture used in the experiments. Repetition number indicates the number of times the block is repeated.}
\label{tab:arch}
\vspace{-.5cm}
\end{table}

\begin{figure}[t!]
\centering
\subfigure[]{\includegraphics[width=0.24\textwidth]{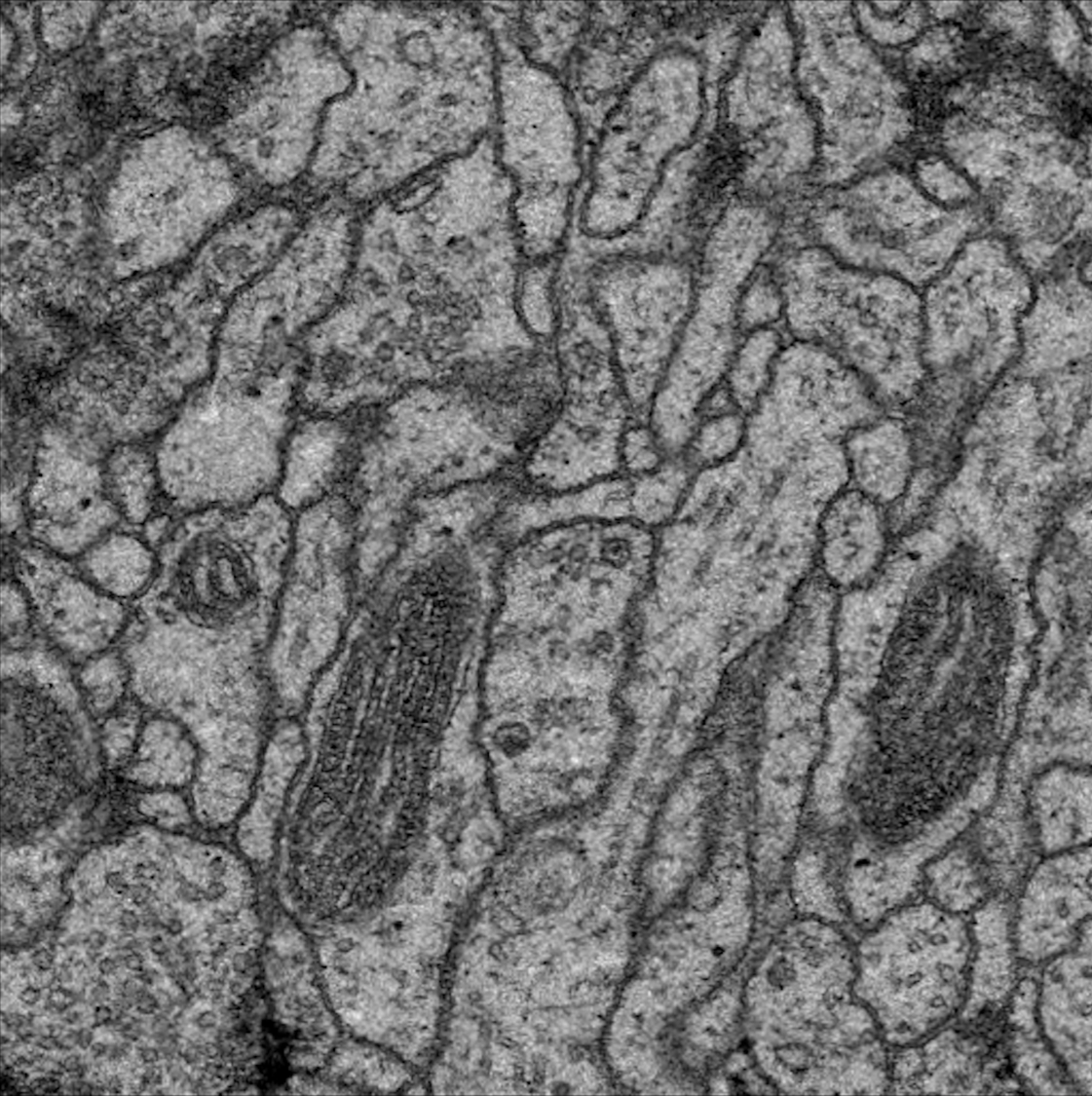}\label{fig:img}}\hfill
\subfigure[]{\includegraphics[width=0.24\textwidth]{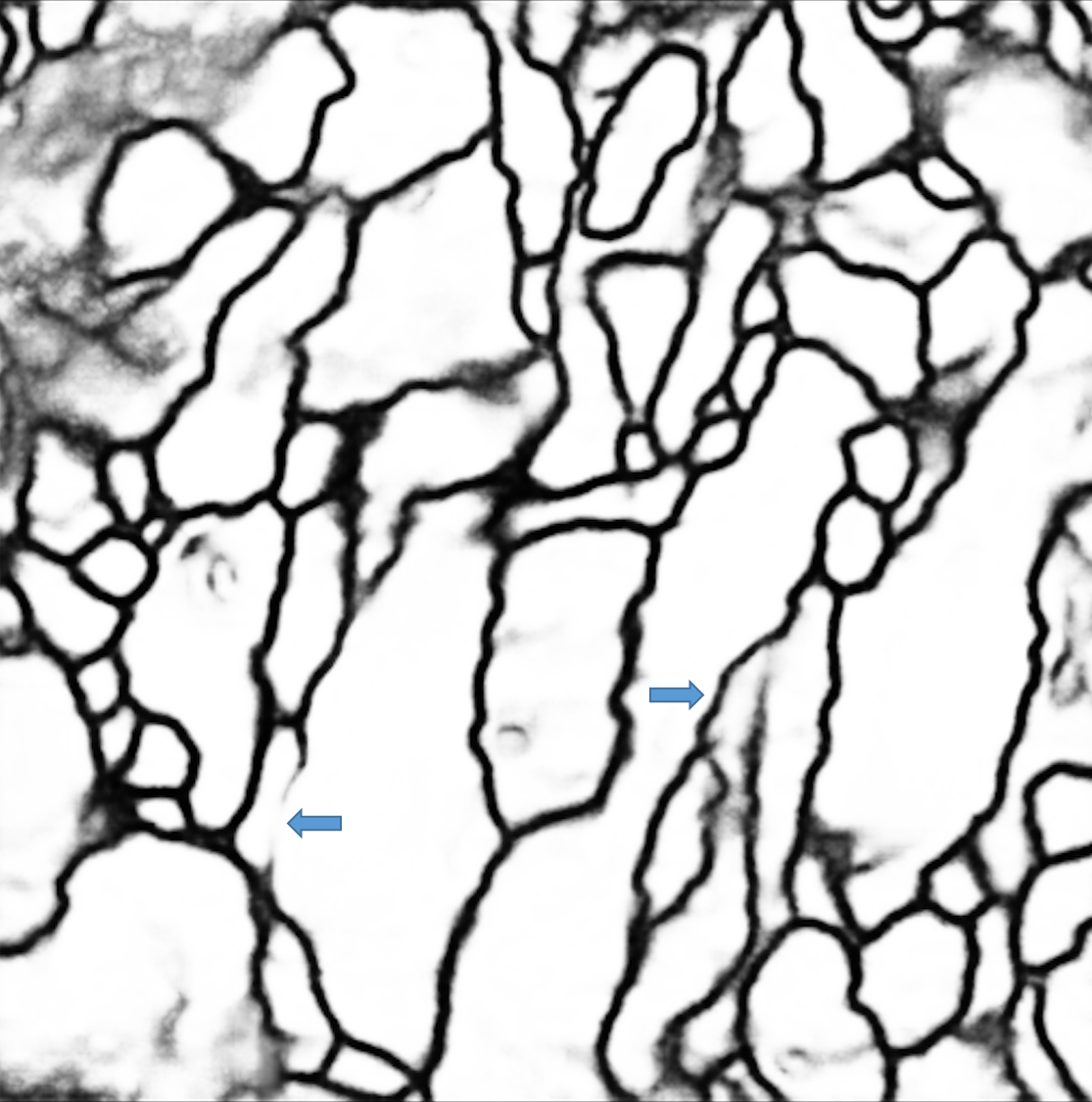}\label{fig:bce}}\hfill
\subfigure[]{\includegraphics[width=0.24\textwidth]{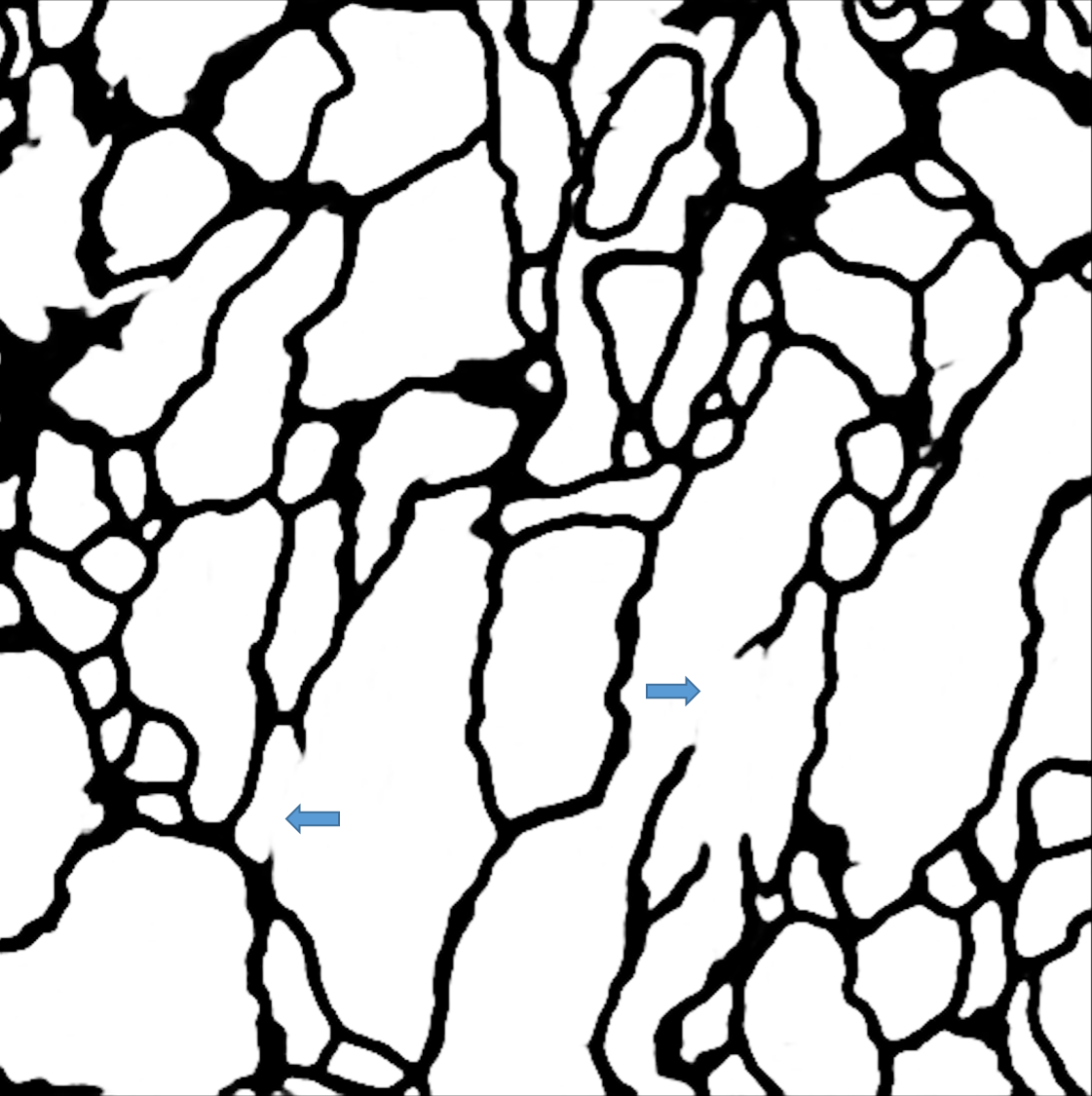}\label{fig:dice}}\hfill
\subfigure[]{\includegraphics[width=0.24\textwidth]{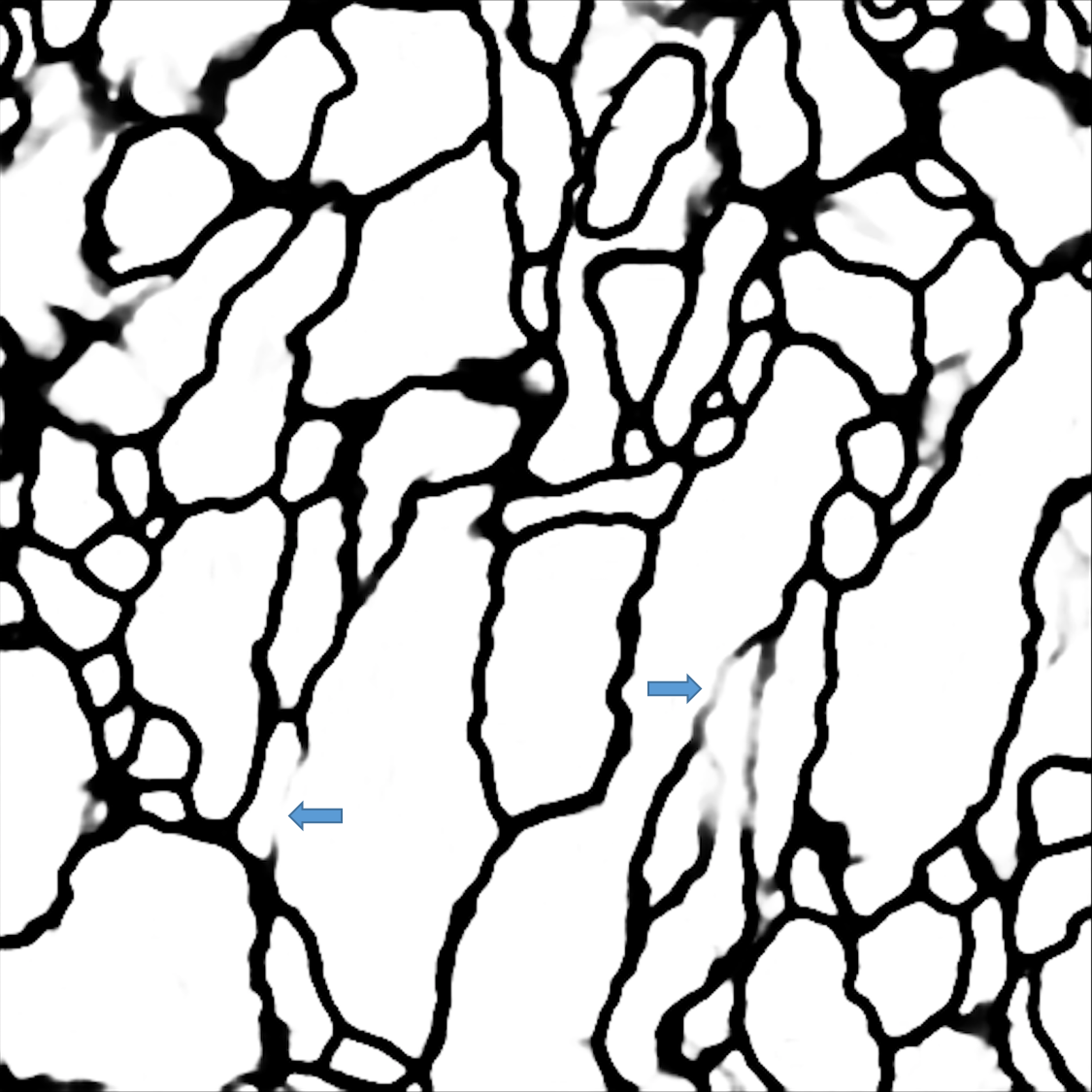}\label{fig:dice2}}\hfill
\vspace{-.1cm}
\caption{Qualitative results on the test set. (a) original image, (b) prediction for a model trained with binary cross-entropy, (c) prediction of the model trained with dice loss and (d) model trained with dice loss with $0.2$ dropout at the test time.}
\label{fig:predictions}
\vspace{-.1cm}
\end{figure}

Interestingly, we found that while the predictions from models trained with cross-entropy loss were of high quality, those produced by models trained with the Dice loss appeared visually cleaner since they were almost binary (similar observations were reported in a parallel work \cite{DBLP:MilletariNA16}.); borders that would appear fuzzy in the former (see Figure \ref{fig:bce}) would be left as gaps in the latter (Figure \ref{fig:dice}). However, we found that the border continuity can be improved for models with the Dice loss by implicit model averaging over output samples drawn at test time, using dropout \cite{KendallBC15} (Figure \ref{fig:dice2}). This yields better performance on the validation and test metrics than the output of models trained with binary cross-entropy (see Table \ref{tab:score}).
%
%

Two metrics used in this dataset are: Maximal foreground-restricted Rand score after thinning ($V_{rand}$) and maximal foreground-restricted information theoretic score after thinning ($V_{info}$). For a detailed description of the metrics, please refer to \cite{EM_data}.

\begin{table}[t!]
\begin{center}
\footnotesize
\begin{tabular}{||c |c |c |c ||c |c |c||} 
\hline
Method & $V_{rand}$ & $V_{info}$ & FCN & post-processing & average over & parameters (M)\\ [0.5ex] 
\hline\hline
CUMedVision \cite{0011QCH16} & 0.977 & 0.989 & YES & YES & 6  &  8 \\
\hline
Unet \cite{RonnebergerFB15} & 0.973 & 0.987 & YES & NO & 7 & 33  \\ 
\hline
IDSIA \cite{Ciresan}  & 0.970 & 0.985 & NO & - & - & -\\
\hline
motif \cite{Wu15f}  & 0.972 & 0.985 & NO & - & - & -\\
\hline
SCI \cite{Liu201488}  & 0.971 & 0.982 & NO & - & - & -\\
\hline
optree-idsia\cite{Uzunbaş2014}  & 0.970 & 0.985 & NO & - & - & -\\
\hline
PyraMiD-LSTM\cite{StollengaBLS15}  & 0.968 & 0.983 & NO & - & - & -\\
\hline
\hline
Ours ($L_{Dice}$)& 0.969 & 0.986 & YES & NO & Dropout & 11 \\
\hline
Ours ($L_{bce}$) & 0.957 & 0.980 & YES & NO & 1 & 11 \\
\hline
\end{tabular}
\end{center}
\caption{Comparison to published entries for EM dataset. For full ranking of all submitted methods please refer to challenge web page: \protect\url{http://brainiac2.mit.edu/isbi_challenge/leaders-board-new}. We note the number of parameter, the use of post-processing, and the use of model averaging only for FCNs.}
\label{tab:score}
\vspace{-.5cm}
\end{table}

Our results are comparable to other published results that establish the state of the art for the EM dataset (Table \ref{tab:score}). Note that we did not do any post-processing of the resulting segmentations. We match the performance of UNet, for which predictions are averaged over seven rotations of the input images, while using less parameters and without sophisticated class weighting. Note that among other FCN available on the leader board, CUMedVision is using post-processing in order to boost performance.

\subsection{On the importance of skip connections}
\label{sec:skip}

\begin{figure}[t!]
\centering
\subfigure[]{\includegraphics[width=0.32\textwidth]{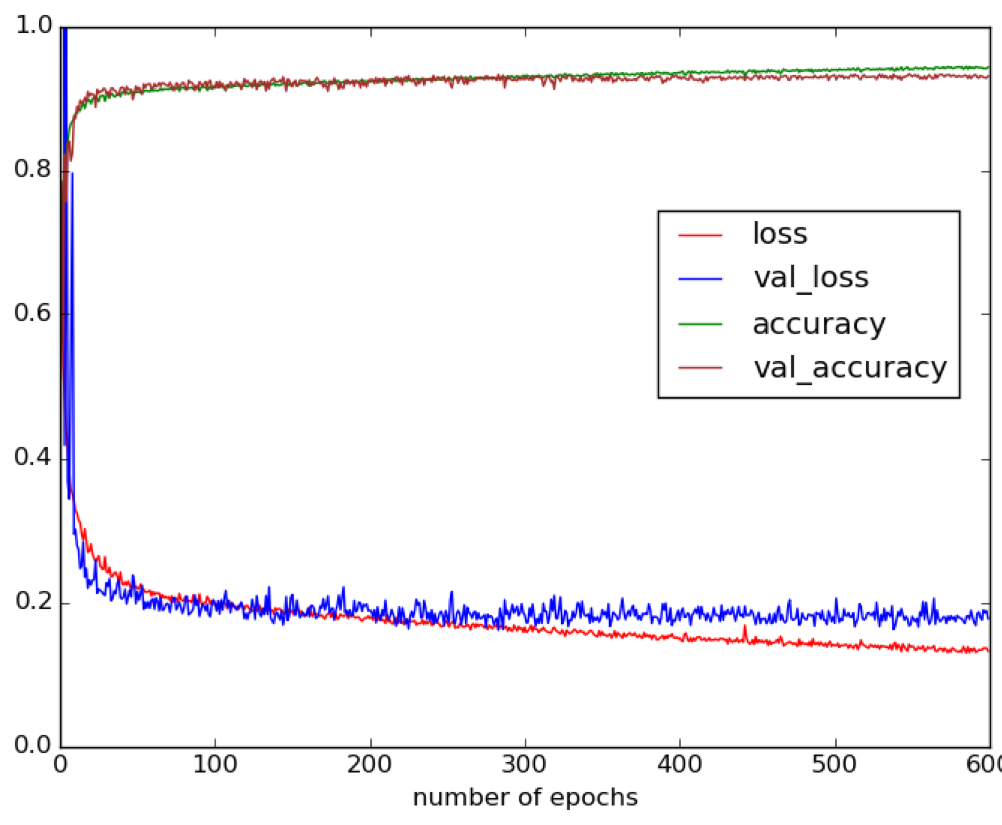}\label{fig:skip1}}\hfill
\subfigure[]{\includegraphics[width=0.32\textwidth]{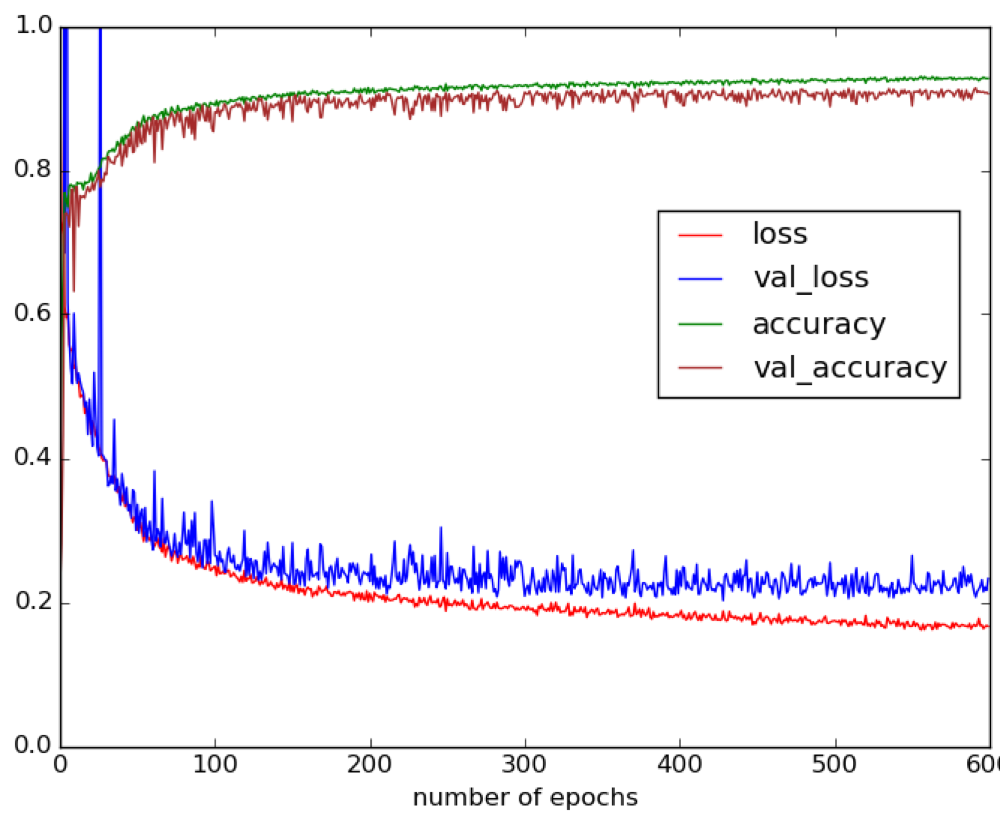}\label{fig:skip2}}\hfill
\subfigure[]{\includegraphics[width=0.32\textwidth]{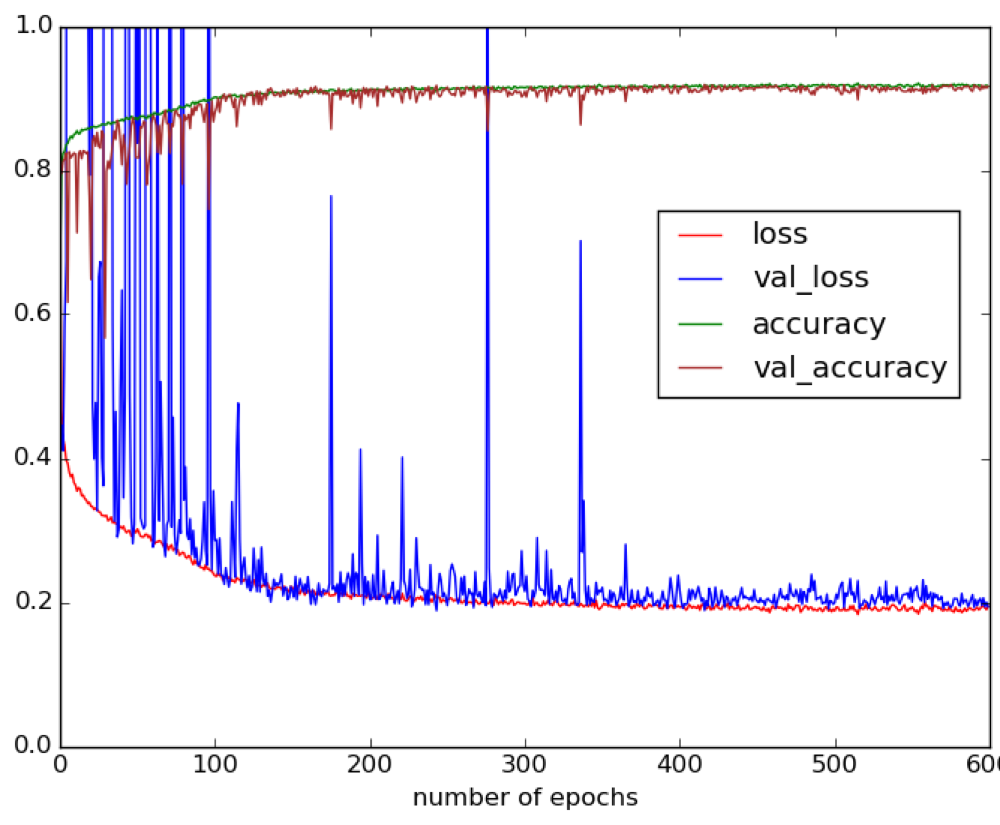}\label{fig:skip3}}\hfill
\vspace{-.2cm}
\caption{Training and validation losses and accuracies for different network setups: (a) Model 1: long and short skip connections enabled, (b) Model 2: only short skip connections enabled and (c) Model 3: only long skip connections enabled.}
\label{fig:skip_conn}
\vspace{-.2cm}
\end{figure}

The focus in the paper is to evaluate the utility of long and short skip connections for training fully convolutional networks for image segmentation. In this section, we investigate the learning behavior of the model with short and with long skip connections, paying specific attention to parameter updates at each layer of the network. We first explored variants of our best performing deep architecture (from Table \ref{tab:arch}), using binary cross-entropy loss. Maintaining the same hyperparameters, we trained (Model 1) with long and short skip connections, (Model 2) with only short skip connections and (Model 3) with only long skip connections. Training curves are presented in Figure \ref{fig:skip_conn} and the final loss and accuracy values on the training and the validation data are presented in Table \ref{tab:loss}.

We note that for our deep architecture, the variant with both long and short skip connections is not only the one that performs best but also converges faster than without short skip connections. This increase in convergence speed is consistent with the literature \cite{SzegedyIV16}. Not surprisingly, the combination of both long and short skip connections performed better than having only one type of skip connection, both in terms of performance and convergence speed. At this depth, a network could not be trained without any skip connections. Finally, short skip connections appear to stabilize updates (note the smoothness of the validation loss plots in Figures \ref{fig:skip1} and \ref{fig:skip2} as compared to Figure \ref{fig:skip3}).

\begin{table}[t!]
\begin{center}
\footnotesize
\begin{tabular}{||c |c |c||} 
\hline
Method & training loss & validation loss \\ [0.5ex] 
\hline\hline
Long and short skip connections & 0.163 & 0.162  \\
\hline
Only short skip connections & 0.188 & 0.202 \\
\hline
Only long skip connection & 0.205 &  0.188\\ 
\hline
\end{tabular}
\end{center}
\caption{Best validation loss and its corresponding training loss for each model.}
\label{tab:loss}
\vspace{-.5cm}
\end{table}

\begin{figure}[t!]
\centering
\subfigure[]{\includegraphics[width=0.215\textwidth]{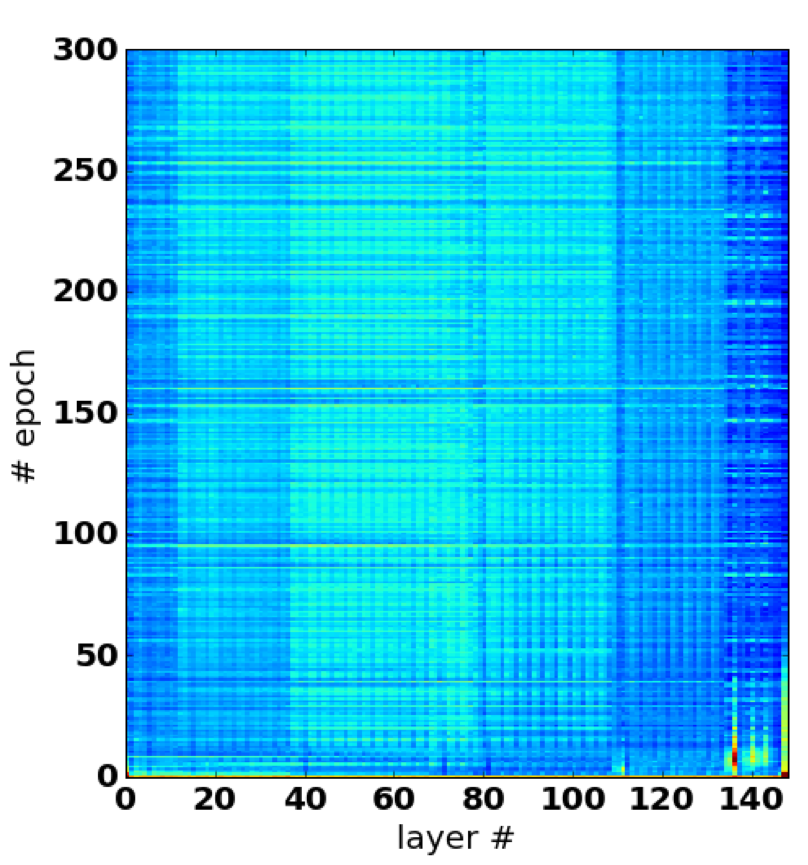}\label{fig:grad1}}\hfill
\subfigure[]{\includegraphics[width=0.21\textwidth]{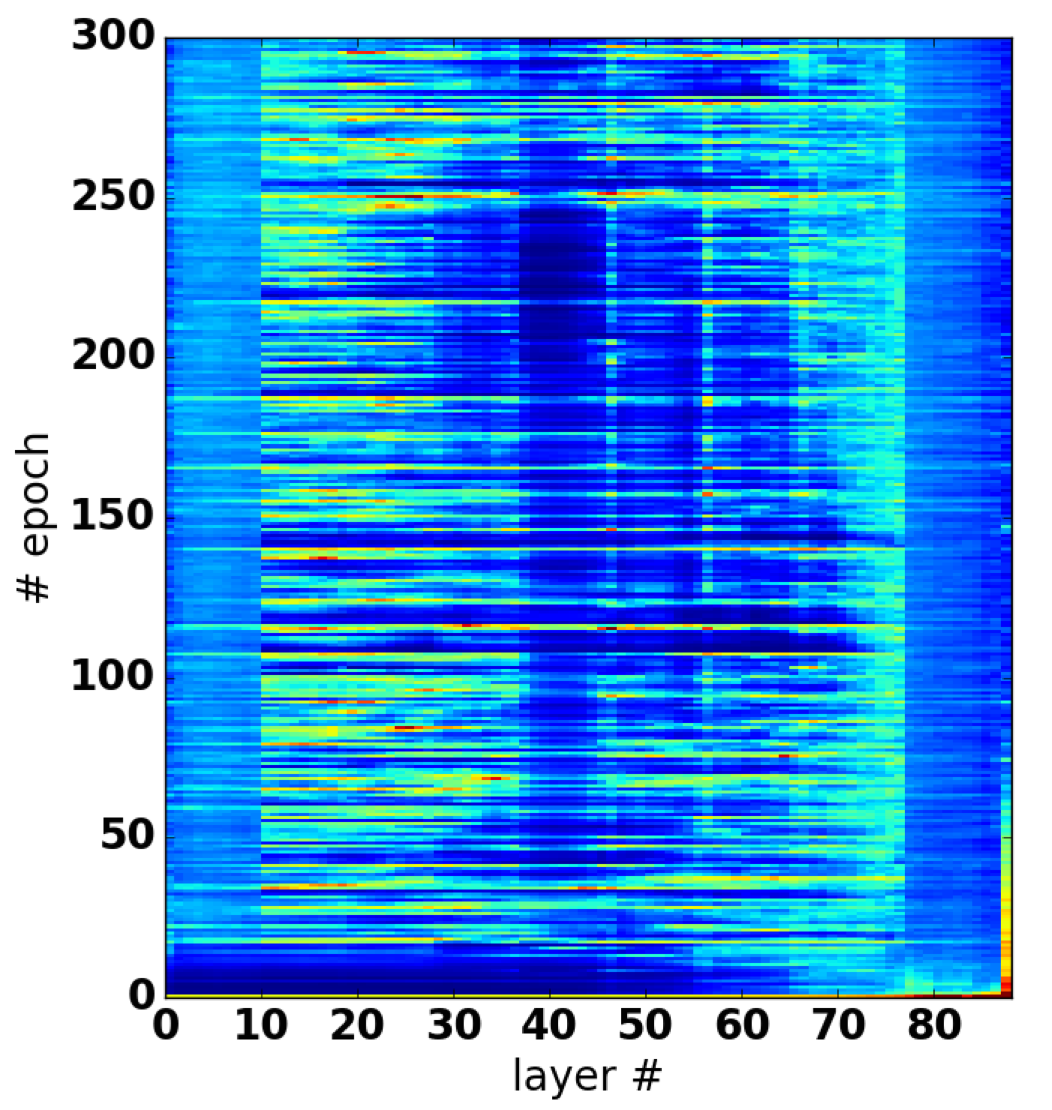}\label{fig:grad2}}\hfill
\subfigure[]{\includegraphics[width=0.255\textwidth]{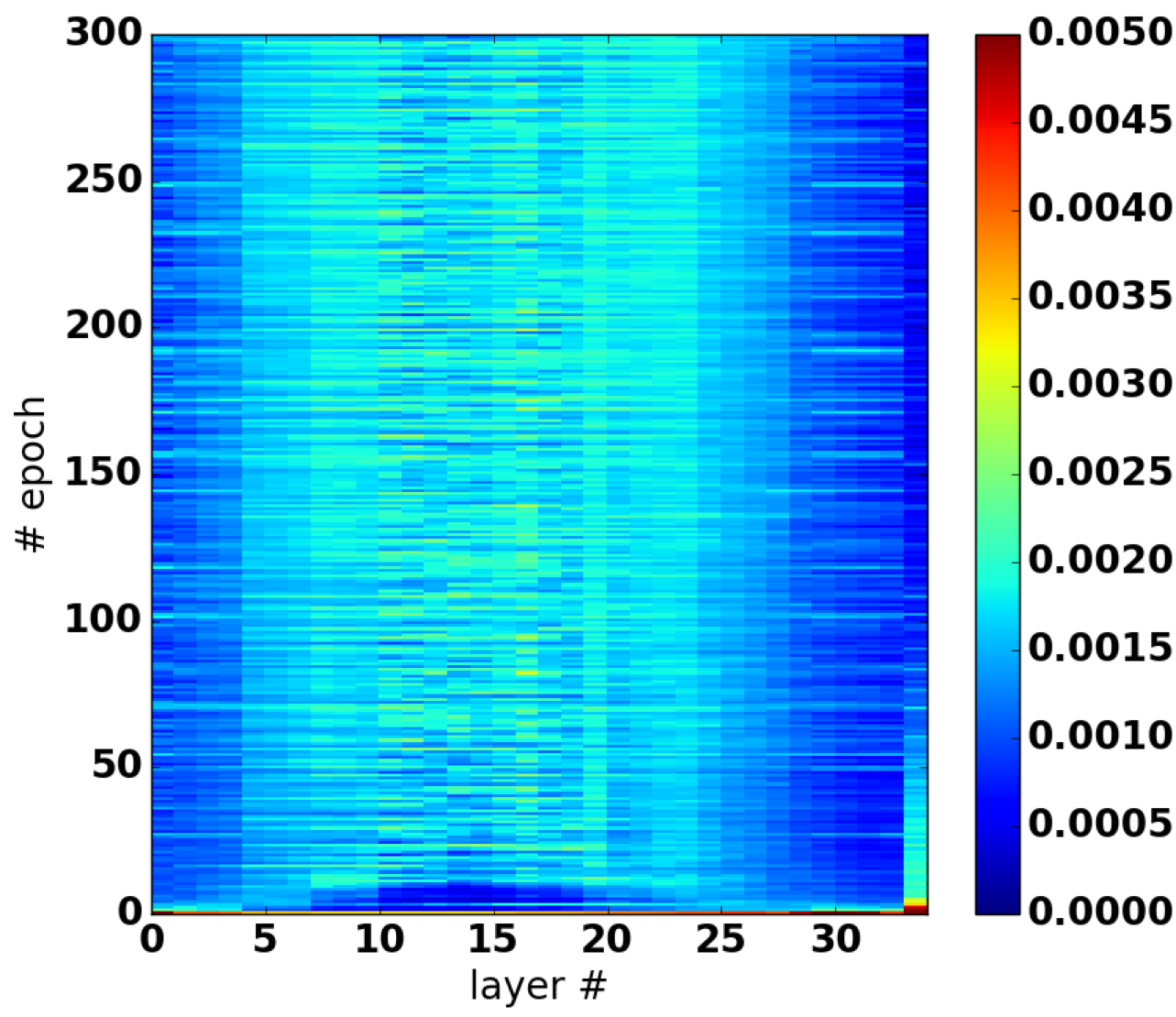}\label{fig:grad3}}\hfill
\subfigure[]{\includegraphics[width=0.275\textwidth]{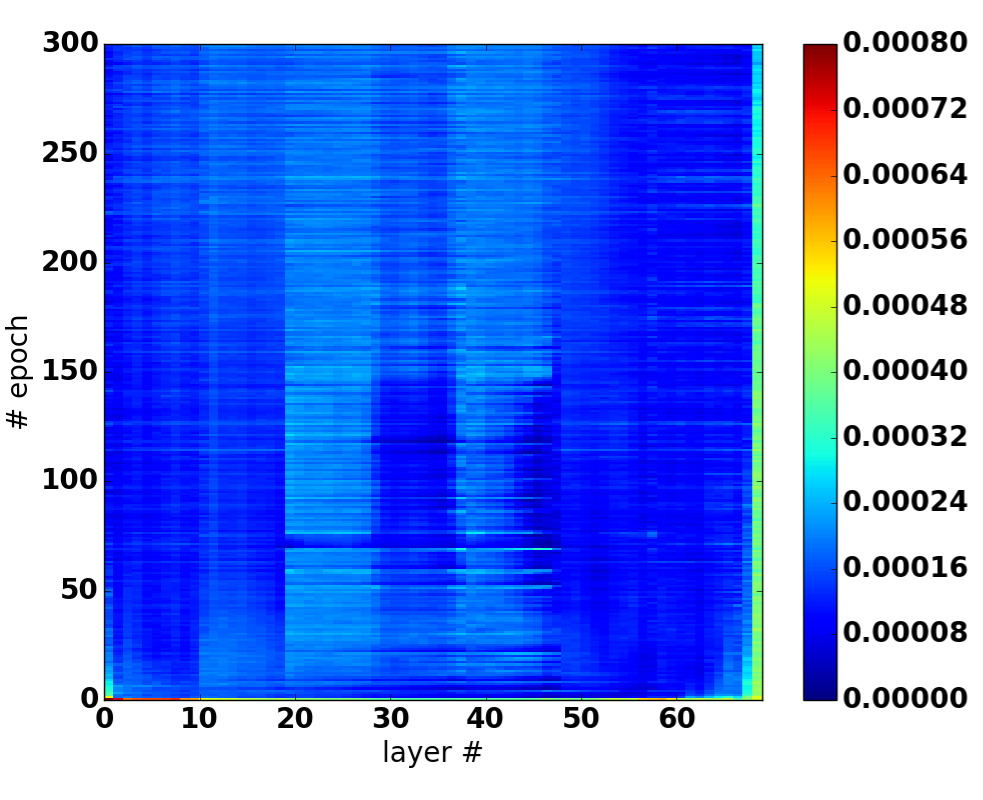}\label{fig:grad4}}\hfill
\vspace{-.2cm}
\caption{Weight updates in different network setups: (a) the best performing model with long and short skip connections enabled, (b) only long skip connections enabled with 9 repetitions of simple block, (c) only long skip connections enabled with 3 repetitions of simple block and (d) only long skip connections enabled with 7 repetitions of simple block, without batch normalization. Note that due to a reduction in the learning rate for Figure (d), the scale is different compared to Figures (a), (b) and (c).}
\label{fig:gradient2}
\vspace{-.5cm}
\end{figure}

We expect that layers closer to the center of the model can not be effectively updated due to the vanishing gradient problem which is alleviated by short skip connections. This identity shortcut effectively introduces shorter paths through fewer non-linearities to the deep layers of our models. We validate this empirically on a range of models of varying depth by visualizing the mean model parameter updates at each layer for each epoch (see sample results in Figure \ref{fig:gradient2}). To simplify the analysis and visualization, we used simple blocks instead of bottleneck blocks.

Parameter updates appear to be well distributed when short skip connections are present (Figure \ref{fig:grad1}). When the short skip connections are removed, we find that for deep models, the deep parts of the network (at the center, Figure \ref{fig:grad2}) get few updates, as expected. When long skip connections are retained, at least the shallow parts of the model can be updated (see both sides of Figure \ref{fig:grad2}) as these connections provide shortcuts for gradient flow. Interestingly, we observed that model performance actually drops when using short skip connections in those models that are shallow enough for all layers to be well updated (eg. Figure \ref{fig:grad3}). Moreover, batch normalization was observed to increase the maximal updatable depth of the network. Networks without batch normalization had diminishing updates toward the center of the network and with long skip connections were less stable, requiring a lower learning rate (eg. Figure \ref{fig:grad4}).

It is also interesting to observe that the bulk of updates in all tested model variations (also visible in those shown in Figure \ref{fig:gradient2}) were always initially near or at the classification layer. This follows the findings of \cite{Saxe_551}, where it is shown that even randomly initialized weights can confer a surprisingly large portion of a model's performance after training only the classifier.

\section{Conclusions}
\label{sec:conclusions}
In this paper, we studied the influence of skip connections on FCN for biomedical image segmentation. We showed that a very deep network can achieve results near the state of the art on the EM dataset without any further post-processing. We confirm that although long skip connections provide a shortcut for gradient flow in shallow layers, they do not alleviate the vanishing gradient problem in deep networks. Consequently, we apply short skip connections to FCNs and confirm that this increases convergence speed and allows training of very deep networks.

\section*{Acknowledgements}
We would like to thank all the developers of Theano and Keras for providing such powerful frameworks. We gratefully acknowledge NVIDIA for GPU donation to our lab at \'{E}cole Polytechnique. The authors would like to thank Lisa di Jorio, Adriana Romero and Nicolas Chapados for insightful discussions. This work was partially funded by Imagia Inc., MITACS (grant number IT05356) and MEDTEQ. 
%

%
%
\bibliographystyle{splncs03}
\bibliography{bib}

\end{document}